\documentclass[10pt,twocolumn,letterpaper]{article}

\usepackage[pagenumbers]{cvpr} %

\usepackage[dvipsnames]{xcolor}

\usepackage{siunitx}
\DeclareSIUnit{\nothing}{\relax}

\usepackage{makecell}

\newcommand{\ulgreen}[1]{{\color{green}\underline{\color{black}#1}}}
\newcommand{\ulyellow}[1]{{\color{yellow}\underline{\color{black}#1}}}
\newcommand{\ulred}[1]{{\color{red}\underline{\color{black}#1}}}

\definecolor{cvprblue}{rgb}{0.21,0.49,0.74}
\usepackage[pagebackref,breaklinks,colorlinks,citecolor=cvprblue]{hyperref}

\def\paperID{*****} %
\def\confName{CVPR}
\def\confYear{2024}

\title{DDOS: The Drone Depth and Obstacle Segmentation Dataset}

\author{Benedikt Kolbeinsson\\
Imperial College London\\
{\tt\small bk915@imperial.ac.uk}
\and
Krystian Mikolajczyk\\
Imperial College London\\
{\tt\small k.mikolajczyk@imperial.ac.uk}
\and
{
\small
\url{https://huggingface.co/datasets/benediktkol/DDOS}}
}

\begin{document}
\maketitle

\begin{abstract}
The advancement of autonomous drones, essential for sectors such as remote sensing and emergency services, is hindered by the absence of training datasets that fully capture the environmental challenges present in real-world scenarios, particularly operations in non-optimal weather conditions and the detection of thin structures like wires. We present the Drone Depth and Obstacle Segmentation (DDOS) dataset to fill this critical gap with a collection of synthetic aerial images, created to provide comprehensive training samples for semantic segmentation and depth estimation. Specifically designed to enhance the identification of thin structures, DDOS allows drones to navigate a wide range of weather conditions, significantly elevating drone training and operational safety. Additionally, this work introduces innovative drone-specific metrics aimed at refining the evaluation of algorithms in depth estimation, with a focus on thin structure detection. These contributions not only pave the way for substantial improvements in autonomous drone technology but also set a new benchmark for future research, opening avenues for further advancements in drone navigation and safety.

\end{abstract}
    
\section{Introduction}
\label{sec:intro}
Fully autonomous drones are poised to revolutionize a multitude of sectors, including remote sensing \cite{tang_drone_2015,inoue2020satellite,mohd2018remote,kellner2019new,bansod2017comparision,shah2023systematic}, package delivery \cite{benarbia2021literature,garg2023drones}, emergency services, and disaster response \cite{erdelj_help_2017,adams2011survey,estrada2019uses,pi2020convolutional,daud2022applications,qu2023environmentally}. 
While manually controlled drones have been effectively employed in specific sectors, the advent of fully autonomous drones is poised to unlock an array of novel applications, enhancing efficiency and expanding capabilities.
However, realizing this potential is contingent upon the ability of drones to navigate safely and autonomously, which in turn requires a precise understanding of their environment. 
Current datasets for training drone navigation systems are inadequate, particularly in representing challenging scenarios such as the detection of thin structures like wires and cables, and operation under diverse weather conditions \cite{mittal2020deep}. This deficiency highlights the need for a dataset that provides a comprehensive representation of the environment, enabling accurate semantic segmentation and depth estimation across a wide range of objects and conditions.

To address this gap, we introduce the Drone Depth and Obstacle Segmentation (DDOS) dataset, a novel resource designed to significantly enhance the training of autonomous drones. DDOS stands out for its dual emphasis on depth and semantic segmentation annotations, with a particular focus on the precise identification of thin structures (a critical but often overlooked aspect in existing datasets). By incorporating advanced computer graphics and rendering techniques, DDOS generates synthetic aerial images that mirror the complexity of real-world environments, encompassing a variety of settings and weather conditions ranging from clear skies to adverse weather scenarios such as rain, fog, and snowstorms.

\begin{table*}[htbp!] \centering
\resizebox{\textwidth}{!}{%
\begin{tabular}{@{}lcccccccccccc@{}}\toprule

\textbf{} &
\textbf{USF} &
\textbf{NE-VBWD} &
\textbf{TTPLA} &
\textbf{PIM} &
\textbf{UAVid} &
\textbf{AeroScapes} &
\textbf{Ruralscapes} &
\textbf{Mid-Air} &
\textbf{TartanAir} &
\textbf{SynthWires} &
\textbf{SynDrone} &
\textbf{DDOS}\\

\textbf{} &
\cite{candamo_detection_2009} &
\cite{stambler_detection_2019} &
\cite{abdelfattah2020ttpla} &
\cite{varghese2017power} &
\cite{lyu2020uavid} &
\cite{nigam2018ensemble} &
\cite{marcu2020semantics} &
\cite{fonder2019mid} &
\cite{wang2020tartanair} &
\cite{madaan2017wire} &
\cite{rizzoli2023syndrone} &
(ours)\\

Data type &
\ulgreen{Real} &
\ulgreen{Real} &
\ulgreen{Real} &
\ulgreen{Real} &
\ulgreen{Real} &
\ulgreen{Real} &
\ulgreen{Real} &
\ulyellow{Synthetic} &
\ulyellow{Synthetic} &
\ulyellow{Synthetic} &
\ulyellow{Synthetic} &
\ulyellow{Synthetic} \\ \midrule

Flight Trajectories  &
\ulred{\num{86}} &
\ulred{\num{41}} &
\ulred{\num{80}} &
\ulred{NA} &
\ulred{\num{30}} &
\ulyellow{\num{141}} &
\ulred{\num{20}} &
\ulred{\num{54}} &
\ulgreen{\num{1037}} &
\ulyellow{\num{154}} &
\ulred{\num{8}} &
\ulgreen{\num{340}}\\

Frames &
\ulyellow{\SI{6}{\kilo\nothing}} &
\ulgreen{\SI{15}{\kilo\nothing}} &
\ulred{\SI{1}{\kilo\nothing}} &
\ulred{\num{159}} &
\ulred{\num{300}} &
\ulyellow{\SI{3}{\kilo\nothing}} &
\ulgreen{\SI{51}{\kilo\nothing}} &
\ulgreen{\SI{119}{\kilo\nothing}\textsuperscript{\textdagger}} &
\ulgreen{\SI{1}{\mega\nothing}} &
\ulgreen{\SI{68}{\kilo\nothing}} &
\ulgreen{\SI{72}{\kilo\nothing}} &
\ulgreen{\SI{34}{\kilo\nothing}} \\

Labeled frames &
\ulyellow{\SI{3}{\kilo\nothing}} &
\ulred{91} &
\ulred{\SI{1}{\kilo\nothing}} &
\ulred{\num{159}} &
\ulred{\num{300}} &
\ulyellow{\SI{3}{\kilo\nothing}} &
\ulred{\SI{1}{\kilo\nothing}*} &
\ulgreen{\SI{119}{\kilo\nothing}\textsuperscript{\textdagger}} &
\ulgreen{\SI{1}{\mega\nothing}} &
\ulgreen{\SI{68}{\kilo\nothing}} &
\ulgreen{\SI{72}{\kilo\nothing}} &
\ulgreen{\SI{34}{\kilo\nothing}} \\ \midrule

Resolution &
\ulred{\(640{\mkern0mu\times\mkern0mu}480\)} &
\ulgreen{\(6576{\mkern0mu\times\mkern0mu}4384\)} &
\ulgreen{\(3840{\mkern0mu\times\mkern0mu}2160\)} &
\ulgreen{\(1280{\mkern0mu\times\mkern0mu}960\)} &
\ulgreen{\(3840{\mkern0mu\times\mkern0mu}2160\)} &
\ulgreen{\(1280{\mkern0mu\times\mkern0mu}720\)} &
\ulgreen{\(3840{\mkern0mu\times\mkern0mu}2160\)} &
\ulgreen{\(1382{\mkern0mu\times\mkern0mu}512\)} &
\ulred{\(640{\mkern0mu\times\mkern0mu}480\)} &
\ulred{\(640{\mkern0mu\times\mkern0mu}480\)} &
\ulgreen{\(1920{\mkern0mu\times\mkern0mu}1080\)} &
\ulgreen{\(1280{\mkern0mu\times\mkern0mu}720\)} \\

Frame rate &
\ulgreen{\SI{25}{\hertz}} &
\ulred{\SI{2}{\hertz}} &
\ulgreen{\SI{30}{\hertz}} &
- & %
\ulred{\SI{0.2}{\hertz}} &
- &
\ulgreen{\SI{50}{\hertz}} &
\ulgreen{\SI{25}{\hertz}} &
- & %
- & %
\ulgreen{\SI{25}{\hertz}} &
\ulyellow{\SI{10}{\hertz}} \\

Environment &
\ulyellow{Town} &
\ulgreen{Town/Nature} &
\ulyellow{Pylons} &
\ulyellow{Pylons} &
\ulgreen{Town/Nature} &
\ulgreen{Various} &
\ulgreen{Town/Nature} &
\ulred{Nature} &
\ulgreen{Various} &
\ulgreen{Various} &
\ulyellow{Town} &
\ulgreen{Town/Nature} \\

Camera motion &
\ulred{Handheld} &
\ulyellow{Helicopter} &
\ulgreen{Drone} &
\ulgreen{Drone} &
\ulgreen{Drone} &
\ulgreen{Drone} &
\ulgreen{Drone} &
\ulgreen{Drone} &
\ulyellow{Random} &
\ulgreen{Drone} &
\ulgreen{Drone} &
\ulgreen{Drone} \\

Altitude &
\ulred{\SI{2}{\meter}} &
\ulred{\(+\)\SI{300}{\meter}} &
- &
- &
\ulred{\SI{50}{\meter}} &
\ulgreen{\SIrange[range-units=single,range-phrase=\,--\,]{5}{50}{\meter}} &
- &
- &
- &
- &
\ulyellow{\(20,50,\,\)\SI{80}{\meter}} &
\ulgreen{\SIrange[range-units=single,range-phrase=\,--\,]{1}{25}{\meter}} \\ \midrule

Weather variations &
\ulred{No} &
\ulred{No} &
\ulred{No} &
\ulred{No} &
\ulred{No} &
\ulred{No} &
\ulred{No} &
\ulgreen{Yes} &
\ulred{No} &
\ulred{No} &
\ulred{No} &
\ulgreen{Yes}\\

Camera pose &
\ulred{No} &
\ulred{No} &
\ulred{No} &
\ulred{No} &
\ulred{No} &
\ulred{No} &
\ulred{No} &
\ulgreen{Yes} &
\ulgreen{Yes} &
\ulred{No} &
\ulgreen{Yes} &
\ulgreen{Yes}\\

Optical flow &
\ulred{No} &
\ulred{No} &
\ulred{No} &
\ulred{No} &
\ulred{No} &
\ulred{No} &
\ulred{No} &
\ulred{No} &
\ulgreen{Yes} &
\ulred{No} &
\ulred{No} &
\ulgreen{Yes}\\

Depth map &
\ulred{No} &
\ulyellow{Sparse} &
\ulred{No} &
\ulred{No} &
\ulred{No} &
\ulred{No} &
\ulred{No} &
\ulgreen{Yes} &
\ulgreen{Yes} &
\ulred{No} &
\ulgreen{Yes} &
\ulgreen{Yes}\\

Segmentation &
\ulred{Wires only} &
\ulred{Wires only} &
\ulgreen{Yes} &
\ulred{No} &
\ulgreen{Yes} &
\ulgreen{Yes} &
\ulgreen{Yes} &
\ulgreen{Yes} &
\ulred{No\textsuperscript{\textdaggerdbl}} &
\ulred{Wires only} &
\ulgreen{Yes} &
\ulgreen{Yes}\\

Thin structures &
\ulgreen{Yes} &
\ulgreen{Yes} &
\ulgreen{Yes} &
\ulyellow{Patches} &
\ulred{No} &
\ulgreen{Yes} &
\ulred{No} &
\ulred{No} &
\ulred{No\textsuperscript{\textdaggerdbl}} &
\ulgreen{Yes} &
\ulred{No} &
\ulgreen{Yes}\\

Mesh structures &
\ulred{No} &
\ulred{No} &
\ulyellow{Rough} &
\ulyellow{Patches} &
\ulred{No} &
\ulyellow{Large only} &
\ulred{No} &
\ulred{No} &
\ulred{No\textsuperscript{\textdaggerdbl}} &
\ulred{No} &
\ulred{No} &
\ulgreen{Yes}\\

\bottomrule
\end{tabular}
}
\caption{\textbf{Comparison between our DDOS dataset and related datasets.} *Ruralscapes also includes automatically generated labels for the remaining \(98\%\) of the dataset. 
\textsuperscript{\textdagger}Mid-Air includes additional variations for the same trajectory.
\textsuperscript{\textdaggerdbl}TartanAir does not include labeled segmentation classes (i.e. each object is assigned to a random unlabeled class, with variations of the same object type in different classes). }
\label{tab:dataset_comparison}
\end{table*}

Our objectives with the DDOS dataset are twofold: firstly, to provide a richly annotated resource that reflects the diversity of scenarios encountered by drones, with a particular focus on thin structures and adverse weather conditions. Secondly, to enable the development and evaluation of algorithms that significantly improve the safety, reliability, and operational efficiency of autonomous drones. By achieving these objectives, we aim to bridge the gap in existing datasets and facilitate the advancement of drone technology to meet the demands of real-world applications.

We present a thorough analysis of DDOS which explores key characteristics including class density, flight dynamics, and spatial distribution, providing a granular understanding of its composition and capabilities. 
Through comparative analysis with existing datasets, we highlight DDOS’s contributions such as incorporating numerous thin and ultra-thin structures with accurate depth and segmentation labels, as well as diverse weather conditions.
Furthermore, we propose new drone-specific metrics designed to accurately evaluate class-specific depth estimation performance. These metrics are tailored to reflect the operational realities of drone applications, offering a refined lens through which to assess algorithmic performance and contributing to the broader goal of advancing drone technology and safety.

Finally, we present baseline results obtained by applying state-of-the-art algorithms to the DDOS dataset, establishing a benchmark for future research in thin object detection. We examine the strengths and limitations of current methodologies, particularly highlighting their notable failure to accurately predict the depth of thin structures. This analysis emphasizes significant opportunities for refinement and innovation within this domain.

To summarize, our main contributions are:
\begin{itemize}
    \item \textbf{DDOS Dataset}: We present the Drone Depth and Obstacle Segmentation (DDOS) dataset, a comprehensive resource developed to significantly improve the training of autonomous drones through extensive depth and semantic segmentation annotations, with a special focus on accurately identifying thin structures.
    \item \textbf{Statistical Analysis and Dataset Comparison}: We provide a thorough examination of the DDOS dataset, highlighting its unique attributes such as class distributions, spatial distribution, and flight dynamics. Our analysis is enriched by a detailed comparative study, positioning DDOS in the broader context of existing datasets and underscoring its distinctive value in addressing specific challenges in drone navigation.
    \item \textbf{Drone-Specific Metrics}: Novel drone-specific metrics are introduced, tailored to the nuances of drone applications, particularly in the evaluation of depth accuracy. These metrics offer a refined and specialized framework for assessing algorithmic performance.
    \item \textbf{Baseline Results and Discussion}: We present baseline results from applying state-of-the-art algorithms to the DDOS dataset, establishing benchmarks for thin object detection research. Our discussion identifies a critical shortfall in existing depth estimation methods, emphasizing the need for future advancements.
\end{itemize}

\section{Related Work}
\label{sec:related_work}

The scarcity of high-quality drone datasets hampers autonomous drone training. This section reviews relevant datasets, evaluating their strengths and weaknesses in regards to training autonomous drones. These evaluations are summarized in \Cref{tab:dataset_comparison}.

\subsection{Driving datasets}
The KITTI \cite{geiger2012we,menze2015object}, Cityscapes \cite{Cordts_2016_CVPR}, nuScenes \cite{nuscenes}, and Waymo \cite{sun2020scalability} datasets, essential in computer vision for autonomous driving, fall short in addressing drone-specific requirements. KITTI's concentration on road scenes lacks the aerial views and diverse thin structures crucial for drone navigation. Similarly, Cityscapes, nuScenes, and Waymo fail to capture the unique aerial perspectives and the slender objects like wires and cables vital for drone safety. The absence of these aerial viewpoints and the limited representation of thin structures mean that models trained on these datasets are not fully equipped to meet the challenges of drone-based navigation.

\subsection{Wire detection datasets}
Several datasets have been specifically designed to tackle the challenge of wire detection, given its critical importance for ensuring the safety of low-flying drones.

The USF dataset \cite{candamo_detection_2009} and NE-VBWD \cite{stambler_detection_2019} are pivotal resources dedicated to wire detection, offering a unique perspective on the challenges of identifying thin structures in aerial imagery. The USF dataset, while extensive, is limited by its image quality and the accuracy of its wire annotations, which are not pixel-accurate and often overlook the real-world curvature of wires, instead defining them as straight lines. This simplification fails to capture the complexity of wire shapes in various environments, undermining the dataset's utility for training models to detect thin structures accurately.
NE-VBWD, although a more recent addition, offers pixel-wise annotations and distance information, focusing on long-range wire detection. However, its suitability for drone applications is limited due to its emphasis on wires located at distances more relevant to manned aircraft, thus diminishing its relevance for low-altitude drone operations where proximity to wires is a critical safety concern.

TTPLA \cite{abdelfattah2020ttpla} and PIM \cite{varghese2017power} also contribute to the field by focusing on transmission towers and power lines, with TTPLA utilizing drone imagery but lacking depth information, and PIM providing small image patches for wire detection without offering semantic segmentation. These datasets, while enriching the domain with specific insights into wire and tower detection, similarly fall short in addressing the broad needs of autonomous drone navigation, such as a diverse range of thin structures, depth mapping, and environmental conditions beyond the mere presence of wires.

\subsection{Drone datasets}
UAVid \cite{lyu2020uavid}, AeroScapes \cite{nigam2018ensemble}, and Ruralscapes \cite{marcu2020semantics} serve as general drone datasets. They provide a broader view of urban and rural landscapes from a drone's perspective, including various object classes for semantic segmentation. Despite their wider scope, these datasets still lack sufficient emphasis on thin structures, such as wires, which are crucial for the safe navigation of drones in complex environments.

SynthWires \cite{madaan2017wire} utilizes a different approach by overlaying synthetic wires over real-world images from drones. This method enhances the variety of wire scenarios available for training, although the absence of depth information limits the dataset's applicability for comprehensive 3D navigation and obstacle avoidance training.

In enhancing the dataset landscape for drone navigation research, Mid-Air \cite{fonder2019mid}, TartanAir \cite{wang2020tartanair}, and SynDrone \cite{rizzoli2023syndrone} represent significant contributions as synthetic datasets offering voluminous labeled training samples. These datasets play a pivotal role in simulating a diverse array of flight dynamics and environmental conditions, providing essential assets such as precise depth maps and camera poses critical for the advancement of sophisticated drone navigation algorithms. Despite their value, these datasets exhibit certain limitations that restrict their comprehensive utility in fully leveraging the potential of synthetic data generation.

One notable shortfall is their failure to encapsulate a complete spectrum of flight scenarios, particularly those involving close encounters, aggressive maneuvering, and very low-altitude flying. Such scenarios, while perilous to execute in real-world settings, are quintessential for preparing drones to navigate through complex, unpredictable environments. Synthetic datasets, with their capacity for controlled simulation, are uniquely positioned to safely incorporate these high-risk flight patterns, thereby enriching the training regime without endangering equipment or safety.

Moreover, while synthetic datasets offer the advantage of generating pixel-perfect segmentation and precise depth measurements, especially for thin structures -- attributes unattainable with conventional data collection methods -- they fall short in representing thin structures like wires, cables, and fences. These elements are critical for ensuring the navigational reliability of drones in densely populated or structurally complex areas. The absence of such objects in the datasets underscores a missed opportunity to leverage some of the benefits of synthetic data generation.

Our proposed dataset, DDOS, is designed to surpass the limitations of existing datasets in wire detection and drone navigation. It provides detailed representations of thin structures and a wide array of other entities, incorporating weather variability and extensive drone motion. Its synthetic foundation enables simulations of close encounters with objects, typically unsafe in reality, enhancing the dataset's utility and realism for drone training.

\begin{figure*}[!ht]
\centering

\makebox[0.195\linewidth]{Image}\hfill
\makebox[0.195\linewidth]{Depth}\hfill
\makebox[0.195\linewidth]{Segmentation}\hfill
\makebox[0.195\linewidth]{Flow}\hfill
\makebox[0.195\linewidth]{Surface normals}\\[0.5mm]

\subfloat{\includegraphics[width=0.195\linewidth]{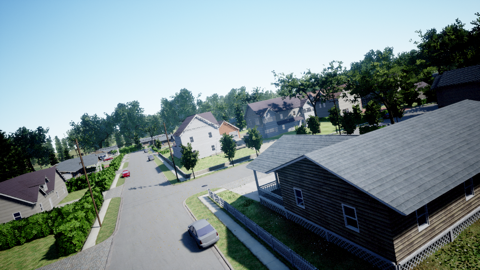}}\hfill
\subfloat{\includegraphics[width=0.195\linewidth]{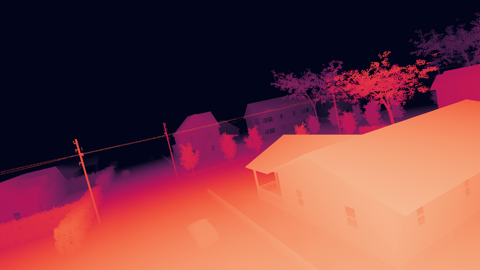}}\hfill
\subfloat{\includegraphics[width=0.195\linewidth]{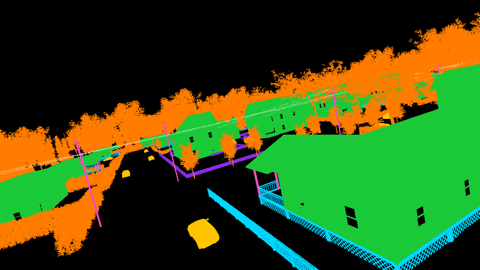}}\hfill
\subfloat{\includegraphics[width=0.195\linewidth]{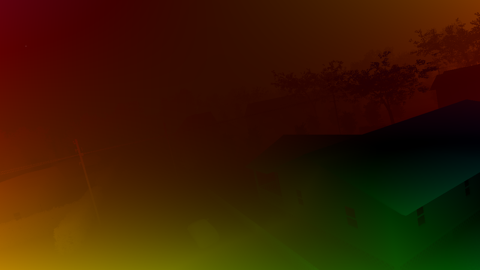}}\hfill
\subfloat{\includegraphics[width=0.195\linewidth]{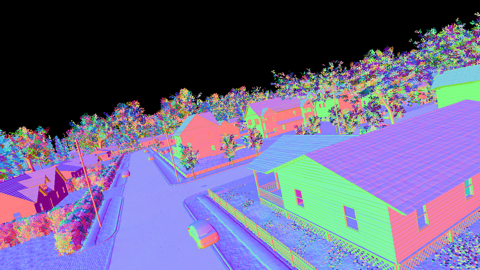}}\\[0.5mm] 

\subfloat{\includegraphics[width=0.195\linewidth]{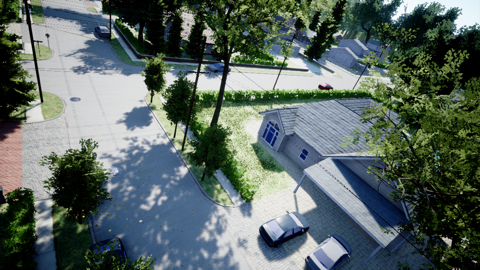}}\hfill
\subfloat{\includegraphics[width=0.195\linewidth]{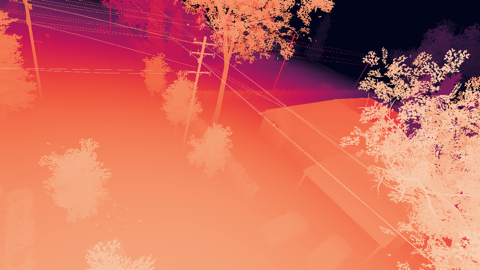}}\hfill
\subfloat{\includegraphics[width=0.195\linewidth]{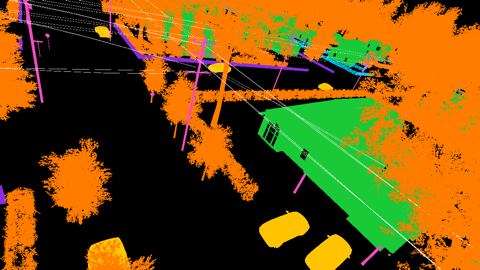}}\hfill
\subfloat{\includegraphics[width=0.195\linewidth]{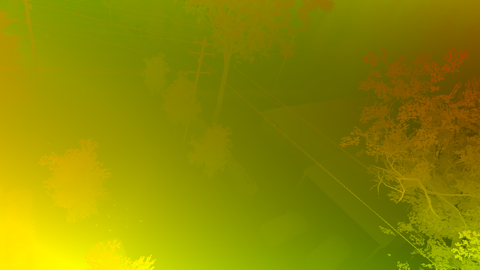}}\hfill
\subfloat{\includegraphics[width=0.195\linewidth]{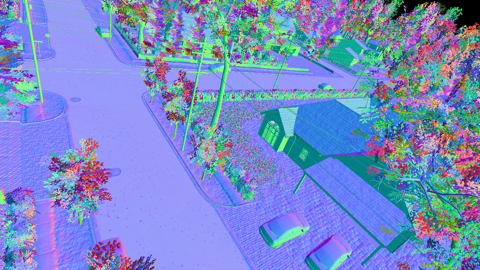}}\\[0.5mm]

\subfloat{\includegraphics[width=0.195\linewidth]{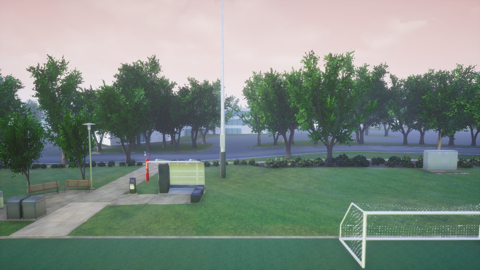}}\hfill
\subfloat{\includegraphics[width=0.195\linewidth]{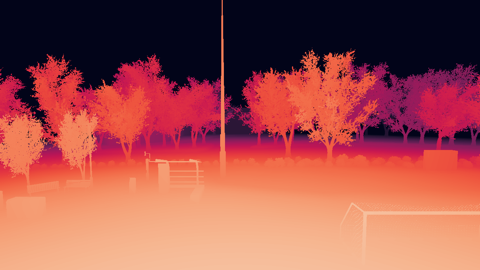}}\hfill
\subfloat{\includegraphics[width=0.195\linewidth]{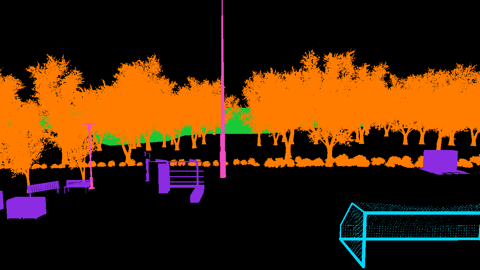}}\hfill
\subfloat{\includegraphics[width=0.195\linewidth]{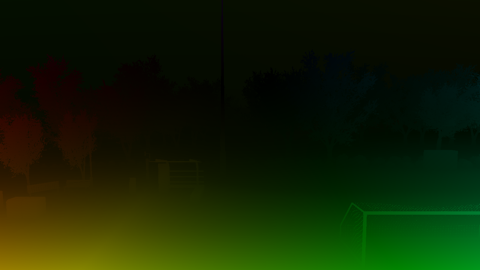}}\hfill
\subfloat{\includegraphics[width=0.195\linewidth]{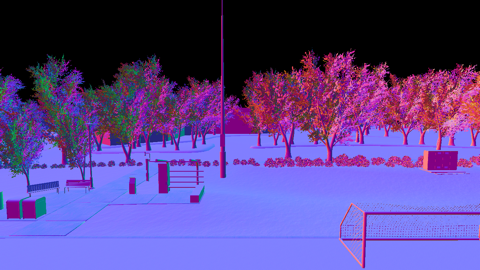}}\\[0.5mm] 

\subfloat{\includegraphics[width=0.195\linewidth]{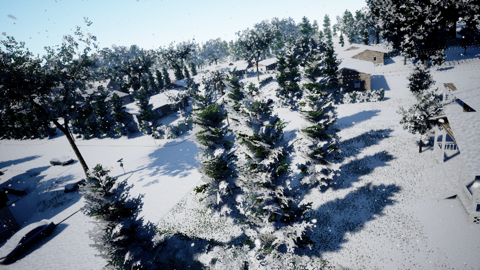}}\hfill
\subfloat{\includegraphics[width=0.195\linewidth]{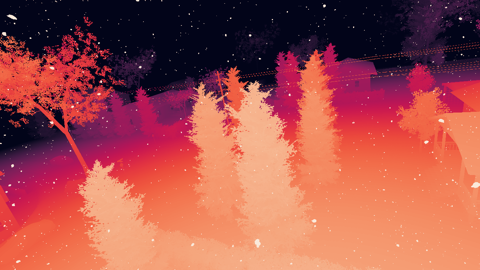}}\hfill
\subfloat{\includegraphics[width=0.195\linewidth]{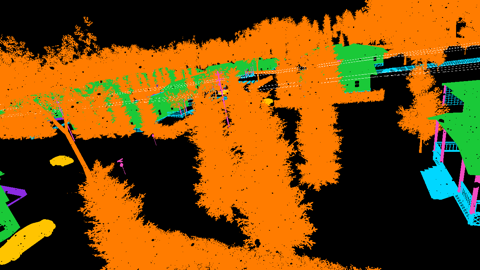}}\hfill
\subfloat{\includegraphics[width=0.195\linewidth]{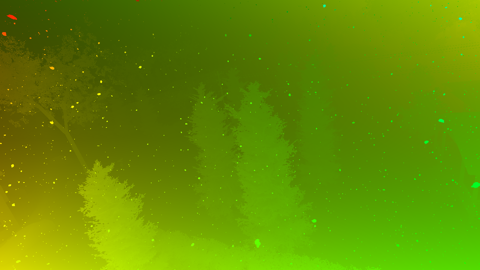}}\hfill
\subfloat{\includegraphics[width=0.195\linewidth]{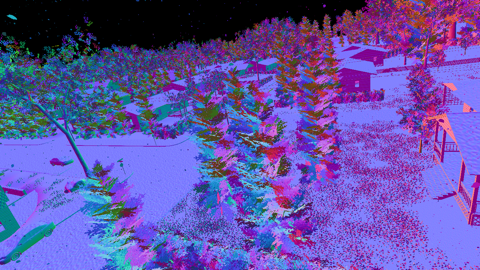}}\\[0.5mm]

\subfloat{\includegraphics[width=0.195\linewidth]{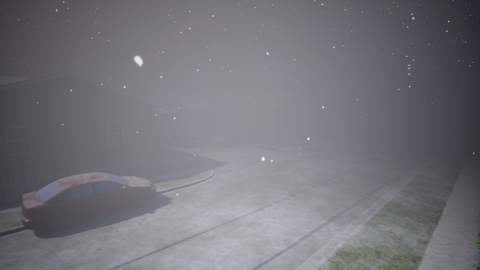}}\hfill
\subfloat{\includegraphics[width=0.195\linewidth]{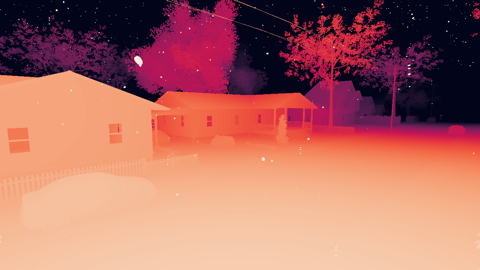}}\hfill
\subfloat{\includegraphics[width=0.195\linewidth]{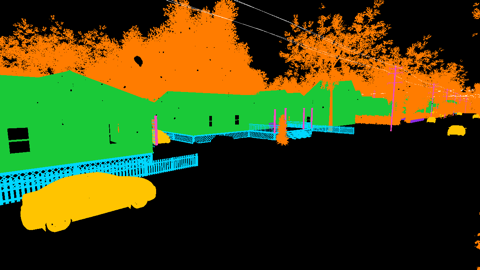}}\hfill
\subfloat{\includegraphics[width=0.195\linewidth]{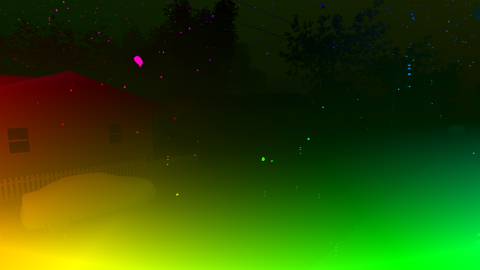}}\hfill
\subfloat{\includegraphics[width=0.195\linewidth]{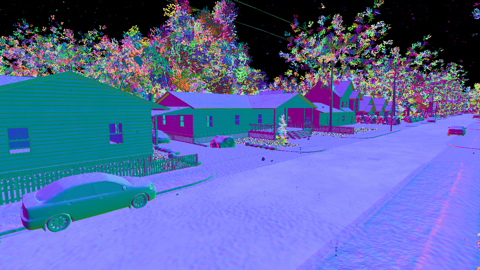}}\\[0.5mm]

\caption{\textbf{Examples from our DDOS dataset.} This figure showcases an overview of the DDOS dataset's multifaceted annotations. It includes RGB images from drone flights, depth maps (\num{0}--\SI{100}{\meter}), pixel-wise semantic segmentation, optical flow and surface normals, illustrating the dataset's richness and diversity.}
\label{fig:ddos_examples}
\end{figure*}

\section{Dataset Features}
\label{sec:dataset_features}

We introduce the DDOS dataset, specifically designed for the training of autonomous drones, utilizing synthetic data generation to compile \num{340} unique drone flights. This dataset is characterized by its comprehensive coverage of various weather conditions, from clear skies to snowstorms, and includes high-risk scenarios such as close encounters and minor collisions. These scenarios, crucial for drone training, are typically too hazardous to replicate in real-world settings. The dataset is notable for its provision of pixel-level precision in semantic segmentation and depth information, particularly for challenging objects such as wires, cables, and fences, thus offering a photo-realistic simulation of environments drones are likely to encounter.

Each flight within the DDOS dataset consists of \num{100} frames, culminating in a total of \num{34000} frames across the dataset. This substantial volume of data supports detailed analysis and algorithm training. The dataset emphasizes thin structures, which present significant navigational challenges, thereby serving as a critical resource for the development of algorithms that require precise segmentation and depth estimation capabilities in complex aerial scenarios. Accompanying the high-resolution images captured by a monocular front facing camera are depth maps, semantic segmentation masks, optical flow data, and surface normals. These components are provided at a resolution of \(1280 \times 720\) pixels, with depth maps covering a range from \SIrange[range-units=single]{0}{100}{\meter}. Additionally, the dataset incorporates exact drone pose, velocity, and acceleration data for each frame.

The DDOS dataset is systematically divided into training, validation, and testing subsets, consisting of \num{300}, \num{20}, and \num{20} flights, respectively. It features pixel-wise segmentation masks for ten distinct classes, enabling in-depth analysis of various obstacles and environmental elements. \Cref{fig:ddos_examples} displays select examples from the dataset, demonstrating the diversity of classes represented. More examples are available in \Cref{sec:additional_examples}. The methodological approach to dataset generation and the classification scheme are further elaborated in \Cref{sec:data_generation}, providing insight into the dataset’s design choices and structure.

\begin{figure*}[!ht]
    \centering
    \includegraphics[width=1.0\linewidth]{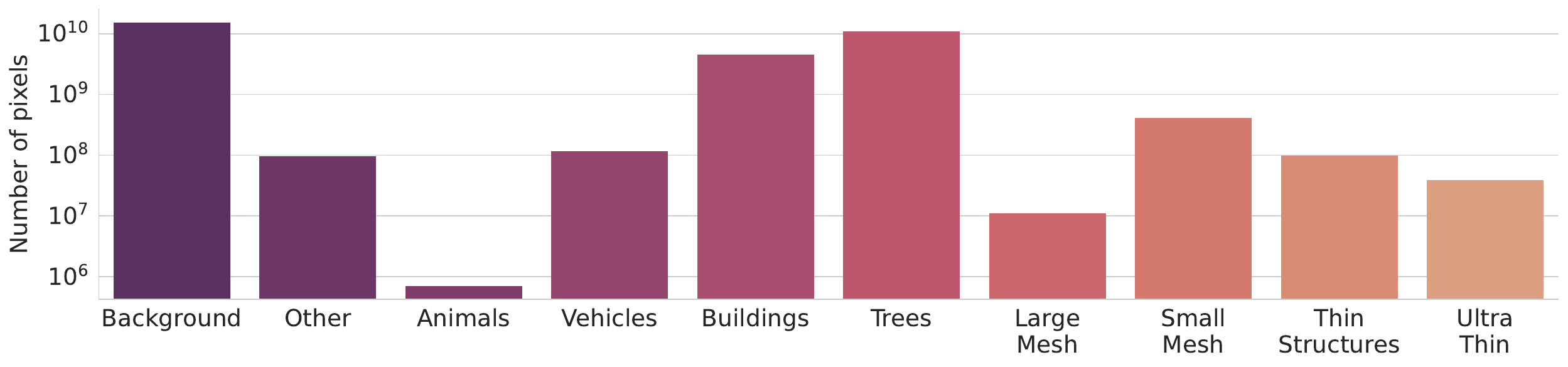}
    \caption{\textbf{Distribution of class labels within DDOS.} DDOS effectively captures the presence of various thin object classes, which are characterized by a relatively sparse distribution of pixels within each image. Despite their limited pixel coverage, these thin object classes are well-represented in DDOS, ensuring comprehensive coverage and enabling robust training and evaluation of algorithms specifically designed to address the challenges posed by such objects.}
    \vspace{0.1em}
    \label{fig:class_distribution}
\end{figure*}

\section{Data Generation}
\label{sec:data_generation}

DDOS is generated using AirSim \cite{airsim2017fsr}, an open-source drone simulator. DDOS is composed of two environments that mimic real-world scenarios. The first environment resembles a small suburban town, featuring dense trees and numerous power lines, replicating the challenges faced during drone flights in residential areas. The second environment represents a park setting, incorporating elements such as a football field with floodlights, a beach volleyball court, dense trees as well as office buildings. These environments collectively offer diverse obstacles and structures, allowing researchers to develop and evaluate algorithms capable of addressing the complexities associated with different real-world environments. By encompassing characteristics like dense tree coverage, power lines, and varying weather conditions, the dataset provides a comprehensive platform for advancing obstacle segmentation and depth estimation algorithms for safe and effective drone flights.

\vspace{-1em}
\paragraph{Flight trajectories}
To construct each flight trajectory, a random starting location \((x_{0}, y_{0}, z_{0})\), within the environment bounds is selected. Subsequently, multiple intermediate target points \((x_{t}, y_{t}, z_{t})\) are generated within predefined relative bounding boxes, dictating the areas to which the drone navigates. Flight characteristics, are varied across different flights, providing diversity in the dataset. During each flight, observations are recorded at a rate of \SI{10}{\hertz} for a duration of \num{10} seconds. These observations encompass a rich set of data, including images, depth maps, pixel-wise object segmentation, optical flow, and surface-normals.

\vspace{-1em}
\paragraph{Collision avoidance}
In order to promote relatively safe flight paths, we developed a dynamic obstacle detection algorithm to modify intermediate targets in response to potential collision risks. This algorithm utilizes the most recent ground truth depth map obtained during the recorded flight observations. By empirically determining a threshold, objects that are deemed too close trigger updates to the intermediate targets. The updated targets are strategically adjusted based on the detected obstacle's location, causing the drone to navigate away from the identified collision risk. This obstacle avoidance approach is not flawless, especially when dealing with thin structures, occasional collisions resulting in crashes still occur. In such cases, the observations associated with the crash event are discarded and the flight process is restarted to ensure data integrity.
It is important to note, the collision avoidance mechanism is purposefully designed to be lax, as near misses and even minor crashes can offer valuable data points for training purposes. 

\vspace{-1em}
\paragraph{Post-processing}
To uphold the overall integrity of the dataset and exclude instances of undesired behavior, additional validation criteria are applied after flight generation. These criteria serve to filter out scenarios where the drone becomes stuck or encounters unusual situations, such as becoming entangled in trees. By incorporating these post-flight validation steps, the dataset ensures that the collected observations reflect reliable and meaningful flight behaviors, enabling robust algorithm training, and evaluation.

\vspace{-1em}
\paragraph{Data augmentation}
We do not augment the dataset with additional transformations or modifications, such as chromatic aberration, added lens flares, corruption, or noise, during the data collection process. The decision to exclude these augmentation techniques at the initial phase ensures that the dataset remains in its original state, preserving the inherent characteristics and properties of the collected data. Instead, we provide the flexibility to incorporate these augmentation techniques at a later stage, if deemed necessary, during algorithm development and evaluation.

\vspace{-1em}
\paragraph{Weather} 
DDOS encompasses diverse environmental and weather conditions, including sunny, dusk, and brightly lit night scenes, along with rain, fog, snow, and changes due to wet surfaces and snow cover. These conditions challenge vision-based algorithms with reduced visibility and altered surface characteristics, such as increased reflectivity from snow and glare from wet roads, complicating object detection and scene analysis. Including these varied scenarios is essential for developing models that adapt and perform consistently in all real-world settings.

\vspace{-1em}
\paragraph{Classes} 
Objects are systematically classified based on their significance for drone navigation. \textit{Ultra Thin} encompasses wires and cables; \textit{Thin Structures} includes poles and signs; \textit{Small Mesh} pertains to fences and nets; and \textit{Large Mesh} covers objects such as transmission towers that permit drone passage. Additionally, \textit{Trees}, \textit{Buildings}, \textit{Vehicles}, and \textit{Animals} are categorized based on straightforward characteristics. The \textit{Other} class encompasses diverse objects like bus stops, post boxes, chairs, and tables. \textit{Background} refers to elements such as the ground and sky, providing context within the scene.

\section{Dataset Statistics}
In this section, we provide a comprehensive analysis of key properties inherent in the DDOS dataset. \Cref{fig:class_distribution} illustrates the distribution of annotations across diverse classes within DDOS. Significantly, the dataset adeptly captures and represents various classes of thin structures, even when these objects occupy a relatively small number of pixels in each image. This nuanced representation ensures that DDOS offers a substantial and well-balanced dataset for thin object classes. This richness in diversity is paramount for facilitating thorough analysis, robust algorithm training, and effective evaluation, particularly in addressing the challenges associated with thin structures in real-world scenarios. The carefully crafted distribution of classes within DDOS contributes to its utility as a reliable benchmark for advancing the capabilities of algorithms designed for thin structure detection and segmentation.

\begin{figure}[t]
    \centering
    \includegraphics[width=1.0\linewidth]{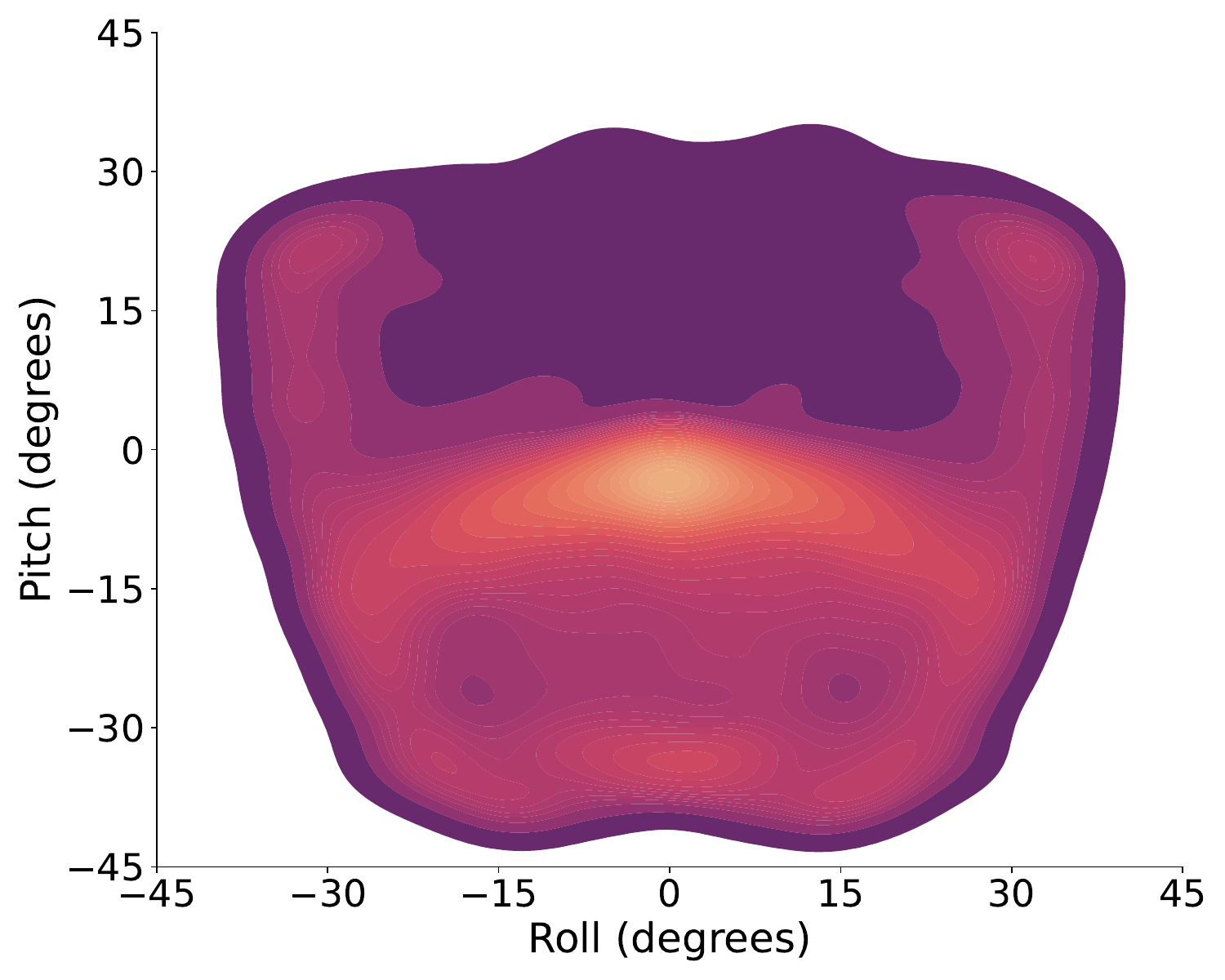}
    \caption{\textbf{Distribution of pitch and roll angles.} The colors represent the intensity levels, with warmer colors indicating higher occurrences. Flight characteristics vary between each flight, as highlighted by the diverse pitch and roll degrees. The pitch is negative when the drone is accelerating forward and positive when braking or to go backwards. Emergency braking is often accompanied with a sharp turn, either to the left or to the right.}
    \label{fig:pitch_roll}
\end{figure}

In our continued investigation, we analyze the pitch and roll angles observed during flight sessions. As depicted in \Cref{fig:pitch_roll}, there is a wide range of pitch and roll angles, indicating significant variations in the drone's orientation across the dataset. Despite the drone's primary forward motion, the angles demonstrate a notable diversity. This variety in orientation provides valuable perspectives for evaluating algorithms under different flight conditions. The broad distribution of pitch and roll angles emphasizes the DDOS dataset's ability to mimic real-world flying scenarios, where drones encounter various orientations. This characteristic enhances the dataset's utility for training and evaluating algorithms to ensure consistent performance amidst the orientation challenges that drones face in actual flights.

\begin{figure}[t]
    \vspace{0.8cm}
    \centering
    \includegraphics[width=1.0\linewidth]{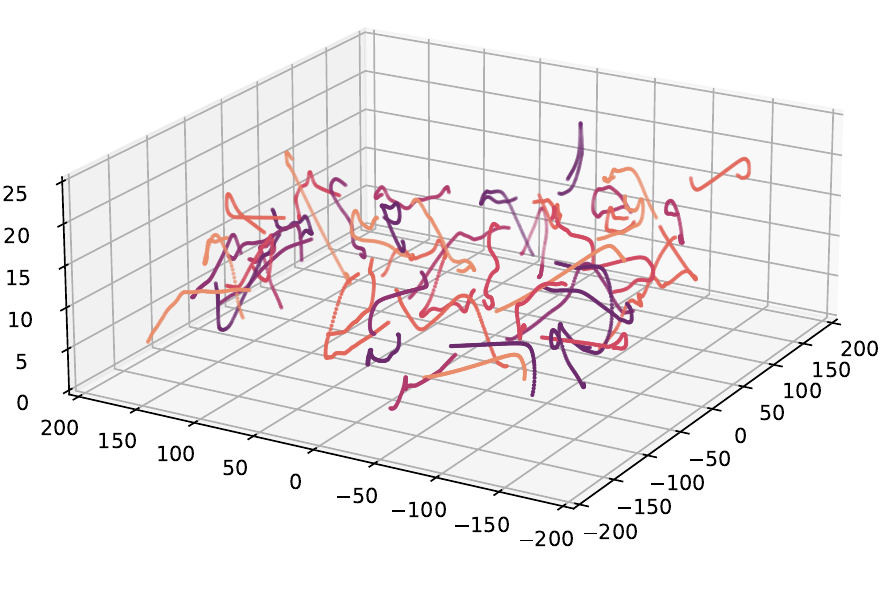}
    \caption{\textbf{Illustrated flight paths.} The figure presents a collection of 50 randomly selected flight paths conducted within the same environment. The paths exhibit significant variations in trajectory, highlighting the diverse nature of drone flights.}
    \label{fig:example_flight_paths}
\end{figure}

\begin{figure}[t]
    \centering
    \includegraphics[width=1.0\linewidth]{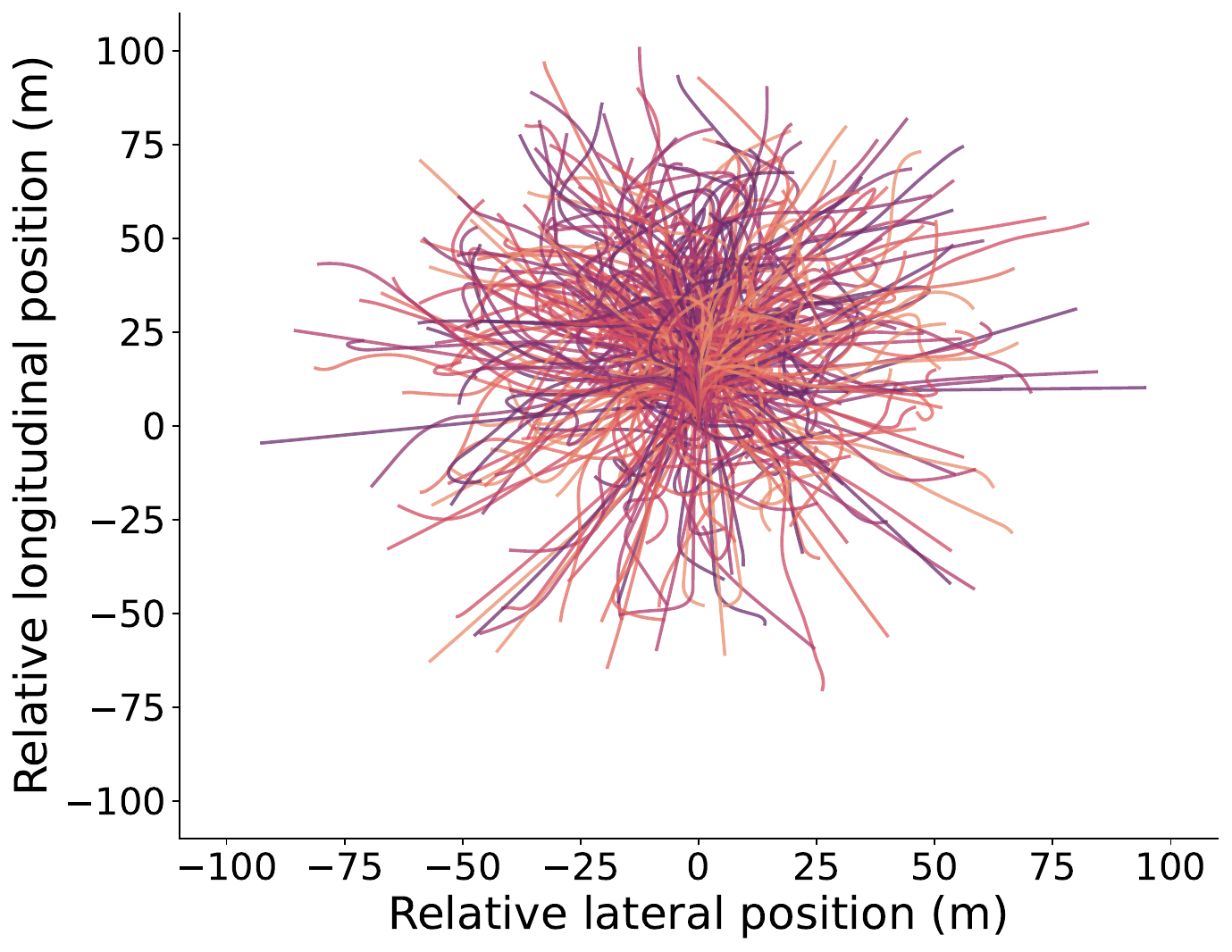}
    \caption{\textbf{Overhead view of relative flight paths with a normalized starting point.} In this visualization the starting location and direction have been normalized to highlight the various relative shapes of the flight paths. The actual starting locations are randomly initialized, as shown in \Cref{fig:example_flight_paths}.}
    \label{fig:overhead_paths}
\end{figure}

To gain an intuitive understanding of the spatial distribution of flight paths within an environment, we visually present a subset of the recorded trajectories in \Cref{fig:example_flight_paths}. The depicted flight paths showcase a diverse array of patterns, ranging from sharp turns and straight lines to curved trajectories. These variations authentically capture the complexity and dynamic nature of the simulated environments. Furthermore, an overhead view of the relative flight paths, presented in \Cref{fig:overhead_paths}, offers a normalized perspective with a common starting point and direction. This visualization emphasizes the diverse flight trajectories and patterns observed across individual flights, providing a comprehensive overview of the spatial dynamics inherent in DDOS. Such a representation is instrumental in offering insights into the intricate navigation challenges that algorithms must address, reinforcing the dataset's efficacy in training and evaluating models under diverse and realistic conditions.

\begin{figure}[t]
  \centering

  \begin{subfigure}[b]{1.0\columnwidth}
    \centering
    \includegraphics[width=\textwidth]{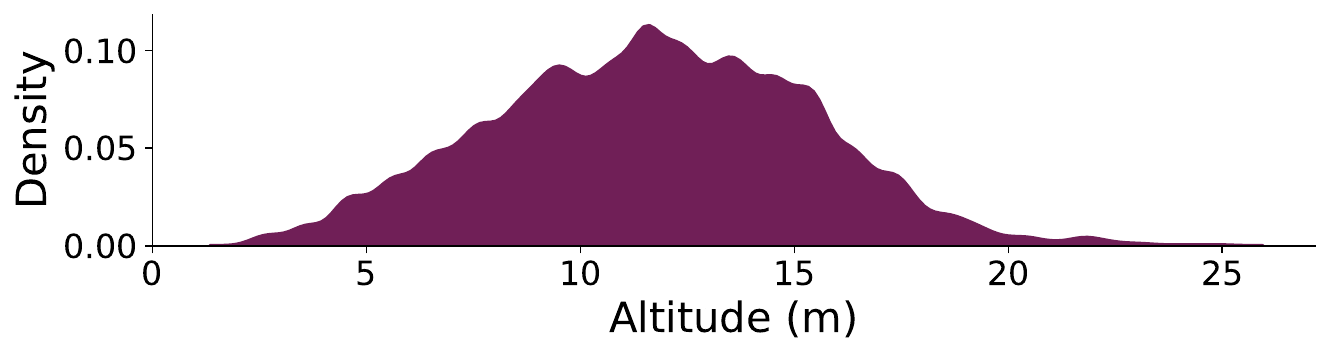}
    \caption{Distribution of flight altitude.}
    \label{fig:distribution_altitude}
  \end{subfigure}
  \hfill
  \begin{subfigure}[b]{1.0\columnwidth}
    \centering
    \includegraphics[width=\textwidth]{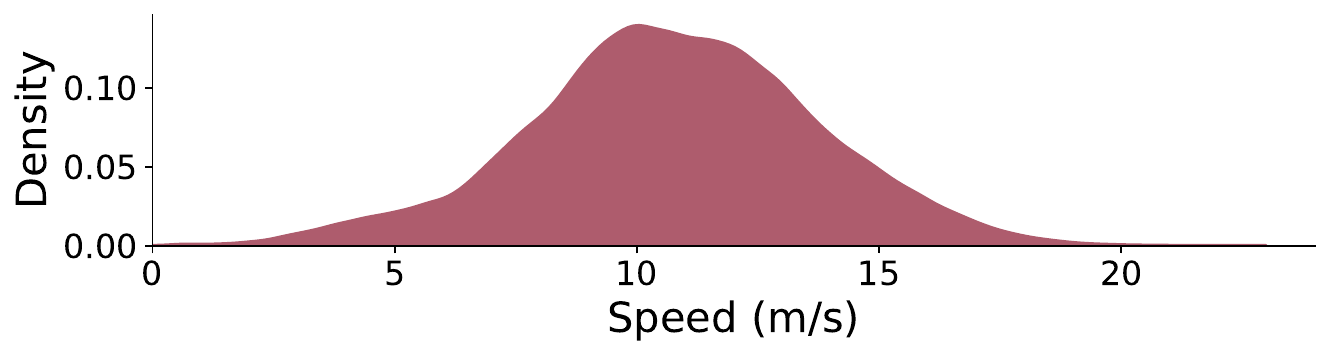}
    \caption{Distribution of flight speed.}
    \label{fig:distribution_speed}
  \end{subfigure}
  \hfill
  \begin{subfigure}[b]{1.0\columnwidth}
    \centering
    \includegraphics[width=\textwidth]{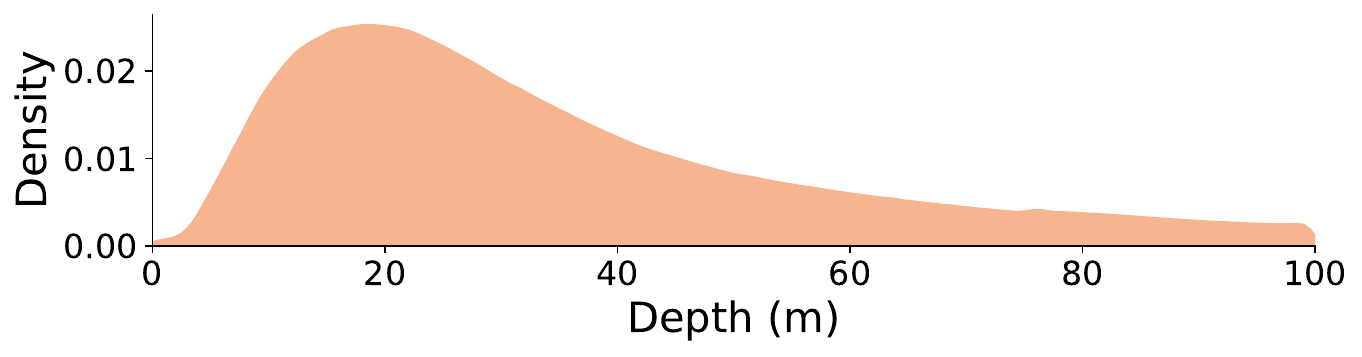}
    \caption{Distribution of depth.}
    \label{fig:distribution_depth}
  \end{subfigure}

  \caption{\textbf{Distributions of altitude, speed, and depth.} The distributions show variation across flights. Depth over \SI{100}{\meter} is ignored.}
  \label{fig:distribution_all}
\end{figure}

Expanding our analysis, we explore the distributions of altitude and speed during the flights, along with the distribution of depth recorded in the depth maps, as illustrated collectively in \Cref{fig:distribution_all}. Examining the altitude distribution reveals that the drone operates at varying heights, encompassing low-level flights near the ground to higher altitudes. The distribution of speed elucidates a spectrum of velocities encountered during the flights, showcasing diverse flight behaviors and maneuvering speeds. Moreover, the depth distribution offers insights into the range and distribution of depth values recorded in the depth maps, shedding light on the variations in perceived depth across the dataset. 

\section{Depth Metrics}
\label{sec:depth_metrics}
We propose a novel set of depth metrics specifically tailored for drone applications, namely the absolute relative depth estimation error for each distinct class. To illustrate, we introduce the absolute relative depth error metric for the \textit{Ultra Thin} class within the DDOS dataset. This metric quantifies the accuracy of depth estimation specifically for objects classified as \textit{Ultra Thin} in the DDOS dataset. 

\begin{equation}
    \text{AbsRel}_{\text{ultra thin}} = \frac{1}{{N_{\text{{ultra thin}}}}} \sum_{{i=1}}^{{N_{\text{{ultra thin}}}}} \left| \frac{{d_i - \hat{d}_i}}{{d_i}} \right|
\end{equation}

Here, \(\text{AbsRel}_{\text{ultra thin}}\) represents the absolute relative depth estimation error for the \textit{Ultra Thin} class. \(N_{\text{{ultra thin}}}\) denotes the total number of samples (pixels) in the \textit{Ultra Thin} class, while \(d_i\) and \(\hat{d}_i\) represent the ground truth depth and estimated depth for the \(i\)-th pixel sample, respectively. The formula calculates the average absolute relative difference between the ground truth and estimated depths for all samples in the \textit{Ultra Thin} class.
Trivially, extending this approach to all classes, the general formula for class-specific depth metrics becomes:

\vspace{-1em}
\begin{equation}
    \text{AbsRel}_{\text{class}} = \frac{1}{{N_{\text{{class}}}}} \sum_{{i=1}}^{{N_{\text{{class}}}}} \left| \frac{{d_i - \hat{d}_i}}{{d_i}} \right|
\end{equation}

Assessing class-specific absolute relative depth errors reveals how well depth estimation algorithms perform, especially for intricate structures like wires and cables. This method offers a detailed evaluation, highlighting how algorithms manage the challenges unique to various structures seen from drone viewpoints. The motivation for this nuanced approach stems from the recognition that traditional metrics fail to adequately represent difficult-to-detect obstacles, such as wires, due to their low pixel count. A thorough investigation into these aspects is essential to accurately gauge the efficacy and robustness of vision systems.

\begin{table*}
  \centering
  \resizebox{\textwidth}{!}{
  \begin{tabular}{lccccccccc}
    \toprule
    Model & \(\delta_1 \uparrow\) & \(\delta_2 \uparrow\) & \(\delta_3 \uparrow\) & AbsRel \(\downarrow\) & RMSE \(\downarrow\) & log10 \(\downarrow\) & RMSElog \(\downarrow\) & SILog \(\downarrow\) & SqRel \(\downarrow\) \\

    \midrule
    BinsFormer~\cite{li2022binsformer} & 0.632 & 0.792 & 0.845 & 0.265 & 16.211 & 0.139 & 0.466 & 38.009 & 6.387 \\
    SimIPU~\cite{li2022simipu} & 0.760 & 0.918 & 0.964 & 0.225 & 7.095 & 0.070 & 0.245 & 22.715 & 3.302 \\
    DepthFormer~\cite{li2023depthformer} & \textbf{0.860} & \textbf{0.958} & \textbf{0.981} & \textbf{0.136} & \textbf{5.831} & \textbf{0.050} & \textbf{0.190} & \textbf{18.101} & \textbf{1.614} \\
    \bottomrule
  \end{tabular}
  }
  \caption{\textbf{Monocular depth estimation performance.} The table compares BinsFormer, SimIPU, and DepthFormer across various traditional performance metrics. Notably, DepthFormer outperforms the other baselines across all metrics, showcasing seemingly great performance in accurately estimating depth. The arrows indicate desired outcome.}
  \label{tab:depth_classic}
\end{table*}

\begin{table*}
  \centering
  \resizebox{\textwidth}{!}{
  \begin{tabular}{lcccccccccc}
    \toprule
    Model & \makecell{Ultra\\Thin} & \makecell{Thin\\Structures} & \makecell{Small\\Mesh} & \makecell{Large\\Mesh} & Trees & Buildings & Vehicles & Animals & Other & Background \\
    \midrule
    BinsFormer~\cite{li2022binsformer} & \textbf{0.945} & \textbf{0.216} & 0.129 & 0.209 & 0.248 & 0.137 & 0.141 & 0.150 & 0.141 & 0.257 \\
    SimIPU~\cite{li2022simipu} & 1.036 & 0.317 & 0.178 & 0.233 & 0.380 & 0.198 & 0.204 & 0.176 & 0.184 & 0.122 \\
    DepthFormer~\cite{li2023depthformer} & 0.998 & 0.229 & \textbf{0.115} & \textbf{0.177} & \textbf{0.206} & \textbf{0.121} & \textbf{0.120} & \textbf{0.121} & \textbf{0.128} & \textbf{0.082} \\
    \bottomrule
  \end{tabular}
  }
  \caption{\textbf{Class-wise absolute relative depth errors.} Each baseline's performance is evaluated per class, with lower values indicating better performance. DepthFormer achieves the lowest errors for the larger classes but completely fails to estimate depth for Ultra Thin. All methods severely struggle for the Ultra Thin class.}
  \label{tab:class_abs_rel}
\end{table*}

\section{Baselines}
\label{baselines}

\begin{figure*}[!ht]
\centering
\makebox[0.195\linewidth]{Input Image}\hfill
\makebox[0.195\linewidth]{Ground Truth}\hfill
\makebox[0.195\linewidth]{BinsFormer~\cite{li2022binsformer}}\hfill
\makebox[0.195\linewidth]{SimIPU~\cite{li2022simipu}}\hfill
\makebox[0.195\linewidth]{DepthFormer~\cite{li2023depthformer}}\\[0.5mm] 
\subfloat{\includegraphics[width=0.195\linewidth]{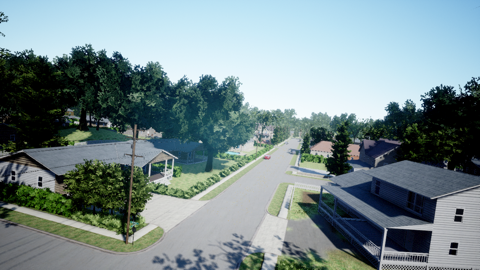}}\hfill
\subfloat{\includegraphics[width=0.195\linewidth]{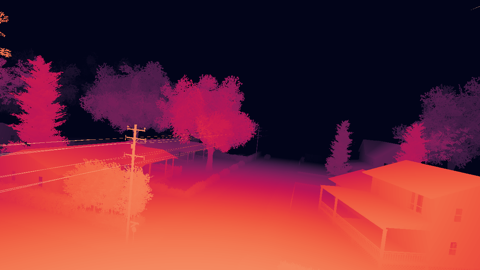}}\hfill
\subfloat{\includegraphics[width=0.195\linewidth]{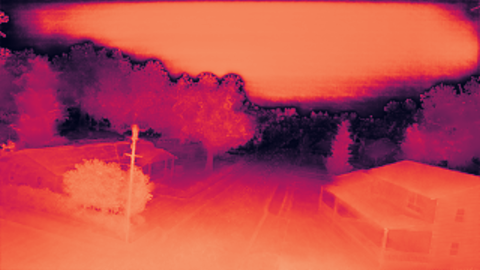}}\hfill
\subfloat{\includegraphics[width=0.195\linewidth]{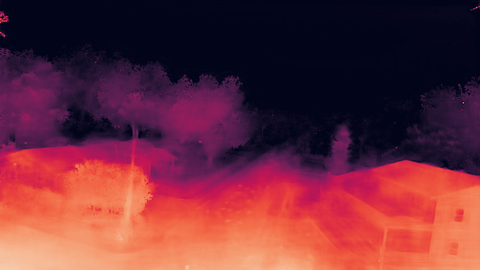}}\hfill
\subfloat{\includegraphics[width=0.195\linewidth]{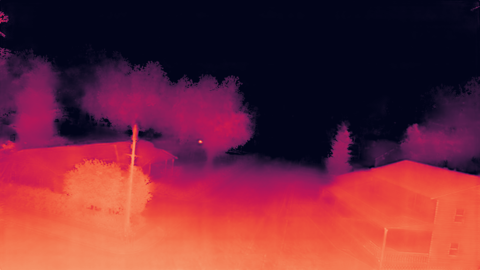}}\\[0.5mm]
\subfloat{\includegraphics[width=0.195\linewidth]{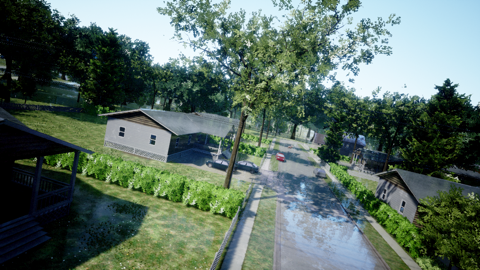}}\hfill
\subfloat{\includegraphics[width=0.195\linewidth]{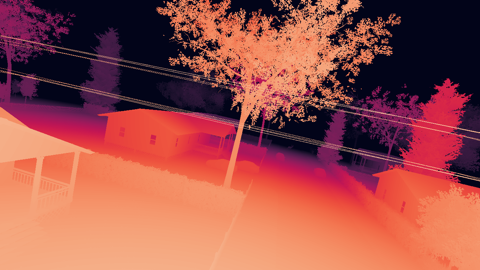}}\hfill
\subfloat{\includegraphics[width=0.195\linewidth]{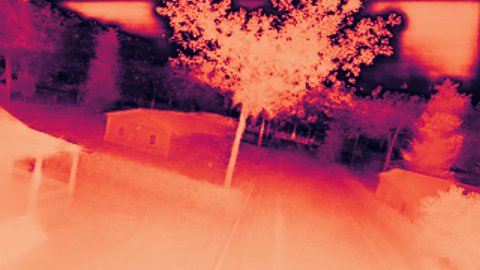}}\hfill
\subfloat{\includegraphics[width=0.195\linewidth]{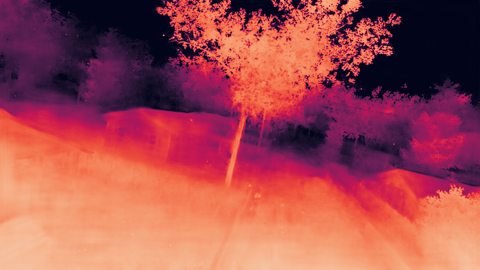}}\hfill
\subfloat{\includegraphics[width=0.195\linewidth]{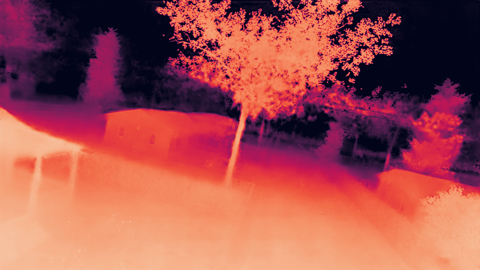}}\\[0.5mm]
\subfloat{\includegraphics[width=0.195\linewidth]{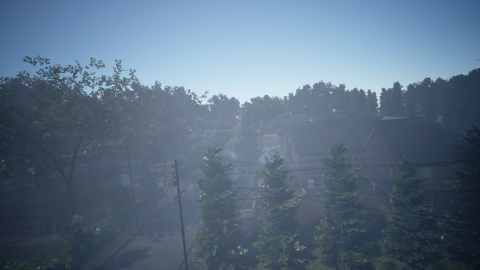}}\hfill
\subfloat{\includegraphics[width=0.195\linewidth]{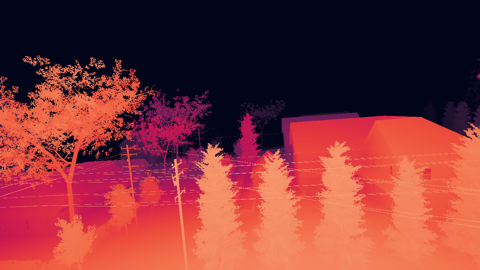}}\hfill
\subfloat{\includegraphics[width=0.195\linewidth]{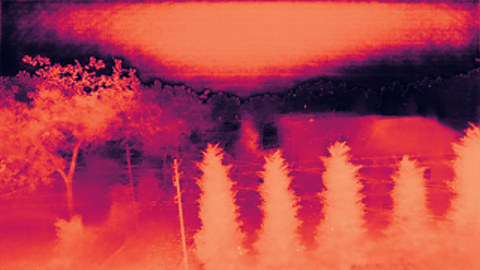}}\hfill
\subfloat{\includegraphics[width=0.195\linewidth]{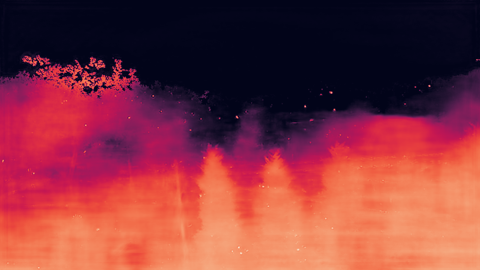}}\hfill
\subfloat{\includegraphics[width=0.195\linewidth]{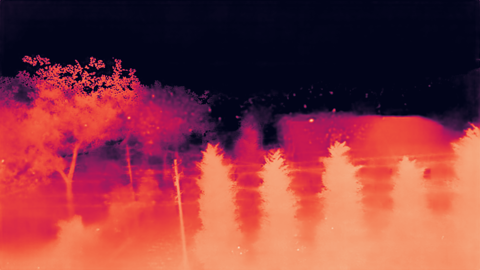}}

\caption{\textbf{Depth estimation performance of baselines.} This qualitative assessment underscores the challenges faced by state-of-the-art methods in accurately estimating depth, particularly for the \textit{Ultra Thin} class. The results showcases the shared difficulty encountered by all methods in capturing the \textit{Ultra Thin} class. This emphasizes the intricate nature of accurately discerning depth for such instances.}
\label{fig:ddos_qualitative}
\end{figure*}

We use a set of commonly-used depth metrics to evaluate the effectiveness of the baselines. These metrics include fundamental measures such as accuracy under the threshold ($\delta_i < 1.25^i$, $i = 1, 2, 3$), which assesses the model's performance within proximity thresholds. Additionally, we use mean absolute relative error (AbsRel), mean squared relative error (SqRel), root mean squared error (RMSE), root mean squared log error (RMSElog), mean log10 error (\text{log10}) and scale-invariant logarithmic error (SILog). 

Moreover, in pursuit of a more nuanced evaluation, we leverage our newly proposed suite of metrics known as mean absolute relative class error metrics (\(AbsRel_{\text{{class}}}\)). This suite is tailored to assess the performance of our methods at a finer class level, offering a more detailed understanding of their capabilities.

We utilize three different baselines, BinsFormer~\cite{li2022binsformer}, SimIPU~\cite{li2022simipu} and DepthFormer~\cite{li2023depthformer}.
BinsFormer proposes a novel framework for monocular depth estimation by formulating it as a classification-regression task, employing a transformer \cite{vaswani2017attention} decoder to generate adaptive bins \cite{bhat2021adabins}. SimIPU introduces a pre-training strategy for spatial-aware visual representation, utilizing point clouds for improved spatial information in contrastive learning. DepthFormer addresses supervised monocular depth estimation by leveraging a transformer for global context modeling, incorporating an additional convolution branch, and introducing a hierarchical aggregation module.

When evaluated using standard depth metrics, the baselines exhibit satisfactory performance, as shown in \Cref{tab:depth_classic}. However, using our class-specific depth metrics, shown in \Cref{tab:class_abs_rel} and depicted in \Cref{fig:ddos_qualitative}, unveils substantial challenges in achieving accurate depth estimations for certain object classes. Specifically, the \textit{Ultra Thin} category is exceptionally challenging, with all tested methods failing to provide accurate depth estimations.

These findings highlight the importance of developing methodologies that are specifically tailored to enhance depth estimation accuracy for ultra-thin structures, particularly in drone-based applications. Future research should focus on addressing these challenges, aiming to enhance the precision and reliability of depth estimations for these challenging scenarios.

\section{Conclusion}

In summary, we introduce the DDOS dataset along with novel drone-specific depth metrics, marking a pivotal advancement in the field of autonomous drone navigation. The DDOS dataset addresses the critical challenges of detecting thin structures and operating under varied weather conditions, thereby filling an essential gap in the current scope of drone research. Through a detailed analysis of the dataset and the deployment of tailored evaluation metrics, we provide a nuanced methodology for systematically assessing the performance of depth estimation algorithms in drone-specific scenarios.

These efforts establish a new standard for future investigations aimed at enhancing the safety and efficiency of drone navigation through superior depth estimation and semantic segmentation techniques. The introduction of the DDOS dataset and corresponding metrics not only propels forward the development of drone technology but also extends the potential for computer vision applications within aerial environments. Our work lays a crucial groundwork for future innovations, steering the creation of algorithms that adeptly navigate the complexities of real-world settings, thus amplifying the functional prowess of drones across a multitude of industries.

\pagebreak

\clearpage
{
    \small
    \bibliographystyle{ieeenat_fullname}
    \bibliography{refs}
}

\maketitlesupplementary

\appendix
\section{Datasheet}
\label{sec:datasheet}
In light of the growing recognition of the pivotal role that datasets play in shaping the behavior and outcomes of machine learning models, this section adheres to the framework proposed in the \textit{Datasheets for Datasets} paper~\cite{gebru2021datasheets}. Acknowledging the potential consequences of mismatches between training or evaluation datasets and real-world deployment contexts, as well as the risk of perpetuating societal biases within machine learning models, we embrace the call for increased transparency and accountability in documenting the provenance, creation, and use of machine learning datasets \cite{wef_human_rights_2018}. By adopting this standardized reporting scheme, we aim to provide a comprehensive understanding of our dataset's motivation, composition, collection process, and recommended uses. This adherence to the datasheets for datasets framework aligns with the broader objective of enhancing transparency, mitigating biases, fostering reproducibility, and aiding researchers and practitioners in selecting datasets tailored to their specific tasks. In the following subsections, we systematically address the key questions outlined in the datasheets for datasets, providing a thorough account of our dataset's characteristics and attributes.

\subsection{Motivation}
\paragraph{For what purpose was the dataset created?}
The Drone Depth and Obstacle Segmentation (DDOS) dataset, was created to address the limitations posed by the scarcity of annotated aerial datasets, specifically for training and evaluating models in depth and semantic segmentation tasks. The primary objective is to focus on the detection and segmentation of thin structures like wires, cables, and fences in aerial views, which are critical for ensuring the safe operation of drones. The dataset aims to fill the gap in existing datasets that predominantly concentrate on common structures and lack representation of fine spatial characteristics of thin structures.

\paragraph{Who created the dataset?}
The dataset was created by Benedikt Kolbeinsson and Krystian Mikolajczyk.

\subsection{Composition}
\paragraph{What do the instances that comprise the dataset represent?}
The instances in the dataset represent individual drone flights which are composed of sequences of observations (images, depth maps, segmentation, etc.) captured during each flight.

\paragraph{How many instances are there in total?}
The dataset consists of a total of 340 drone flights, and each flight comprises \num{100} sequential observations. Therefore, there are a total of \num{34000} observations (\num{340} flights \(\times\) \num{100} observations per flight).

\paragraph{Does the dataset contain all possible instances or is it a sample (not necessarily random) of instances from a larger set?}
No, there exists many more possible flight paths in the environments used as well as in other environments.

\paragraph{What data does each instance consist of?}
Each flight consists of \num{100} sequential observations, comprising of a high-resolution image captured by a monocular camera affixed to the front of the drone, corresponding depth maps, pixel-level object segmentation masks, optical flow information and surface normals. As well as coordinates, pose and speed information and environment information including weather. All image modalities maintain a resolution of \num{1280}\(\times\)\num{720}, and the depth maps cover a range from \SIrange[range-units=single]{0}{100}{\meter}.

\paragraph{Is there a label or target associated with each instance?}
Yes, DDOS features pixel-wise object segmentation masks with ten distinct classes, allowing for detailed analysis of diverse obstacles and environmental elements. These classes are: \textit{ultra thin}, \textit{thin}, \textit{small mesh}, \textit{large mesh}, \textit{trees}, \textit{buildings}, \textit{vehicles}, \textit{animals}, \textit{other}, and \textit{background}. For instance, the \textit{ultra thin} class covers objects like wires and cables, while the \textit{thin} class encompasses streetlights and poles. The \textit{small mesh} class includes objects like fences and nets, and the \textit{large mesh} class involves structures similar to pylons and radio masts.
In addition, corresponding depth maps, optical flow information and surface normals are included.

\paragraph{Is any information missing from individual instances?}
No.

\paragraph{Are relationships between individual instances made explicit?}
Yes, the flight coordinates are available.

\paragraph{Are there recommended data splits?}
Yes, the dataset is partitioned into training, validation, and testing subsets, encompassing 300, 20, and 20 flights, respectively.

\paragraph{Are there any errors, sources of noise, or redundancies in the dataset?}
The data is simulated and no artificial noise is added.

\paragraph{Is the dataset self-contained, or does it link to or otherwise rely on external resources?}
Yes, DDOS is self-contained.

\paragraph{Does the dataset contain data that might be considered confidential?}
No.

\paragraph{Does the dataset contain data that, if viewed directly, might be offensive, insulting, threatening, or might otherwise cause anxiety?}
No.

\subsection{Collection Process}

\paragraph{How was the data associated with each instance acquired?}
The data was acquired through simulated drone flights using AirSim~\cite{airsim2017fsr}, a drone simulator.

\paragraph{What mechanisms or procedures were used to collect the data?}
DDOS was generated using AirSim and data was saved using built-in APIs.

\paragraph{If the dataset is a sample from a larger set, what was the sampling strategy?}
During the simulation process, flights with severe crashes were discarded.

\paragraph{Who was involved in the data collection process?}
Data collection scripts were written by Benedikt Kolbeinsson.

\paragraph{Over what timeframe was the data collected?}
The simulation process took two days.

\paragraph{Were any ethical review processes conducted?}
No.

\subsection{Preprocessing / cleaning / labeling}
\paragraph{Was any preprocessing / cleaning / labeling of the data done?}
During the simulation, labels such as depth and semantic segmentation are automatically recorded. Flights with severe crashes were discarded. 

\paragraph{Was the “raw” data saved in addition to the preprocessed / cleaned / labeled data?}
The processed data is a lossless function of the raw data. The only removed data are flights with severe crashes and are not saved.

\paragraph{Is the software that was used to preprocess / clean / label the data available?}
Yes, AirSim is open source.

\subsection{Uses}
\paragraph{Has the dataset been used for any tasks already?}
No.

\paragraph{Is there a repository that links to any or all papers or systems that use the dataset?}
No.

\paragraph{What (other) tasks could the dataset be used for?}
DDOS is valuable for training and evaluating algorithms related to obstacle and object segmentation, depth estimation, and drone navigation.

\paragraph{Is there anything about the composition of the dataset or the way it was collected and preprocessed / cleaned / labeled that might impact future uses?}
No.

\paragraph{Are there tasks for which the dataset should not be used?}
Yes, DDOS should not be used for malicious purposes.

\subsection{Distribution}
\paragraph{Will the dataset be distributed to third parties outside of the entity on behalf of which the dataset was created?}
Yes, DDOS is hosted on Hugging Face and is available at: 

\noindent
\href{https://huggingface.co/datasets/benediktkol/DDOS}{huggingface.co/datasets/benediktkol/DDOS}

\paragraph{How will the dataset be distributed?}
DDOS is openly available on Hugging Face: 

\noindent
\href{https://huggingface.co/datasets/benediktkol/DDOS}{huggingface.co/datasets/benediktkol/DDOS}

\paragraph{When will the dataset be distributed?}
On publication of this paper.

\paragraph{Will the dataset be distributed under a copyright or other intellectual property (IP) license, and/or under applicable terms of use (ToU)?}
Yes, DDOS is openly licensed under \href{https://creativecommons.org/licenses/by-nc/4.0/}{CC BY-NC 4.0}.

\paragraph{Have any third parties imposed IP-based or other restrictions on the data associated with the instances?}
No.

\paragraph{Do any export controls or other regulatory restrictions apply to the dataset or to individual instances?}
No.

\begin{figure*}[t]
\centering
\makebox[0.328\linewidth]{Image}\hfill
\makebox[0.328\linewidth]{Depth}\hfill
\makebox[0.328\linewidth]{Segmentation}\hfill\\[0.5mm]

\subfloat{\includegraphics[width=0.328\linewidth]{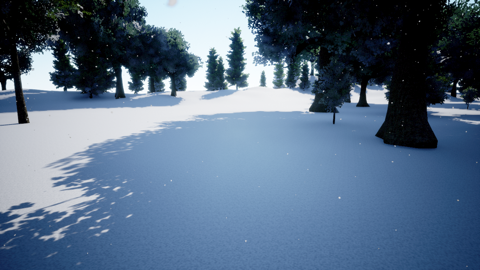}}\hfill
\subfloat{\includegraphics[width=0.328\linewidth]{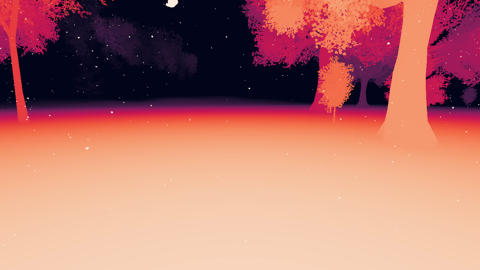}}\hfill
\subfloat{\includegraphics[width=0.328\linewidth]{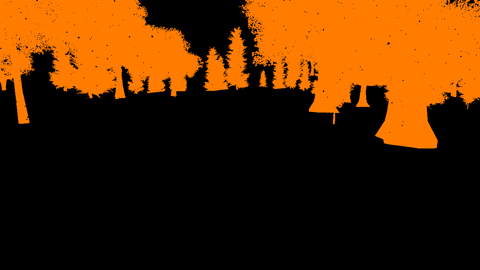}}\\[0.5mm]
\subfloat{\includegraphics[width=0.328\linewidth]{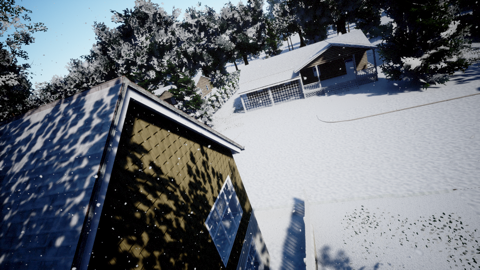}}\hfill
\subfloat{\includegraphics[width=0.328\linewidth]{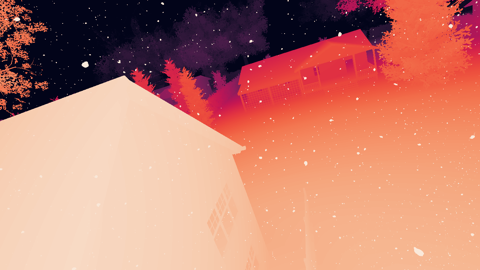}}\hfill
\subfloat{\includegraphics[width=0.328\linewidth]{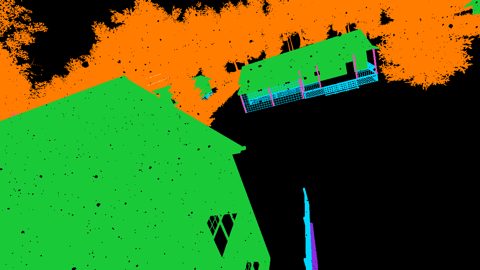}}\\[0.5mm]
\subfloat{\includegraphics[width=0.328\linewidth]{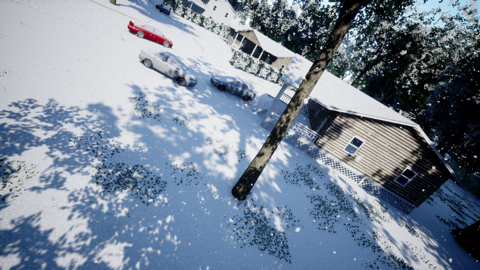}}\hfill
\subfloat{\includegraphics[width=0.328\linewidth]{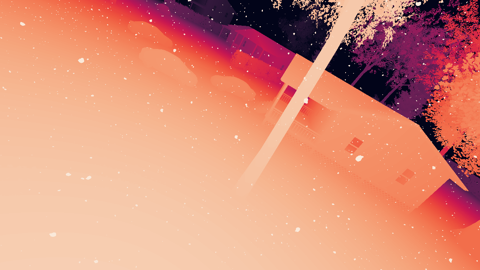}}\hfill
\subfloat{\includegraphics[width=0.328\linewidth]{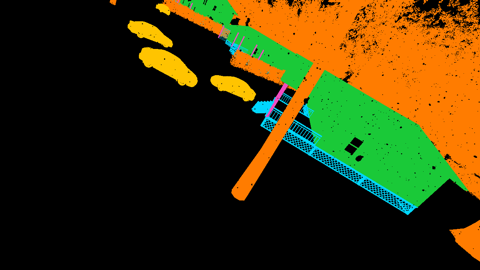}}

\caption{\textbf{Low altitude examples from DDOS.} The DDOS dataset encompasses flights featuring diverse flight characteristics, including examples of low altitude maneuvers and aggressive turns under snowy conditions.}
\label{fig:ddos_additional_examples2}
\end{figure*}

\subsection{Maintenance}
\paragraph{Who will be supporting / hosting / maintaining the dataset?}
DDOS is hosted on Hugging Face

\paragraph{How can the owner / curator / manager of the dataset be contacted?}
Contact can be made on Hugging Face:

\noindent
\href{https://huggingface.co/datasets/benediktkol/DDOS}{huggingface.co/datasets/benediktkol/DDOS}

\paragraph{Is there an erratum?}
No.

\paragraph{Will the dataset be updated?}
There is no current plan to augment the dataset.

\paragraph{If the dataset relates to people, are there applicable limits on the retention of the data associated with the instances?}
Not applicable.

\paragraph{Will older versions of the dataset continue to be supported / hosted / maintained?}
Yes.

\paragraph{If others want to extend / augment / build on / contribute to the dataset, is there a mechanism for them to do so?}
There is no specific mechanism for others to extend / augment / build on / contribute to the dataset.

\begin{figure*}[p]
\centering
\makebox[0.328\linewidth]{Image}\hfill
\makebox[0.328\linewidth]{Depth}\hfill
\makebox[0.328\linewidth]{Segmentation}\hfill\\[0.5mm]

\subfloat{\includegraphics[width=0.328\linewidth]{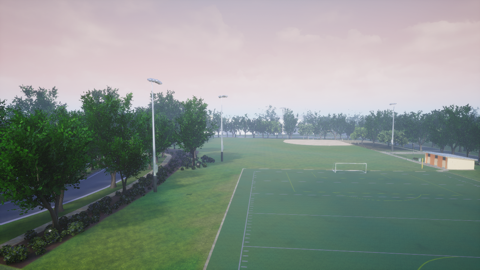}}\hfill
\subfloat{\includegraphics[width=0.328\linewidth]{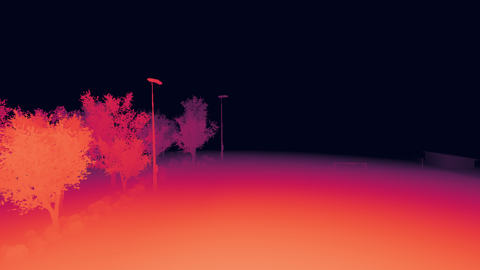}}\hfill
\subfloat{\includegraphics[width=0.328\linewidth]{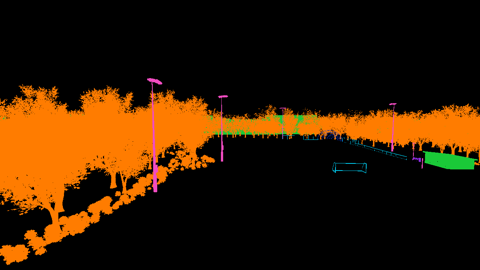}}\\[0.5mm]
\subfloat{\includegraphics[width=0.328\linewidth]{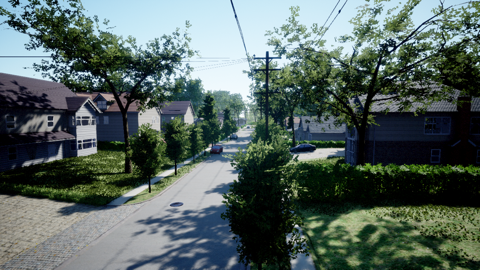}}\hfill
\subfloat{\includegraphics[width=0.328\linewidth]{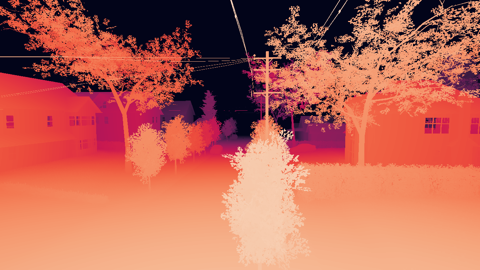}}\hfill
\subfloat{\includegraphics[width=0.328\linewidth]{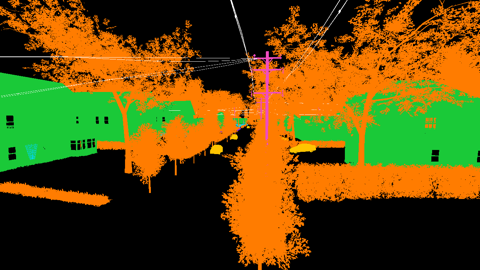}}\\[0.5mm]
\subfloat{\includegraphics[width=0.328\linewidth]{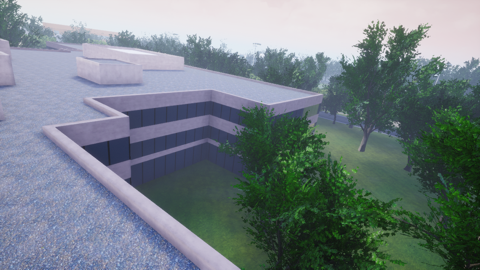}}\hfill
\subfloat{\includegraphics[width=0.328\linewidth]{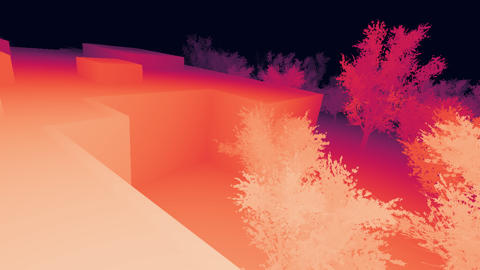}}\hfill
\subfloat{\includegraphics[width=0.328\linewidth]{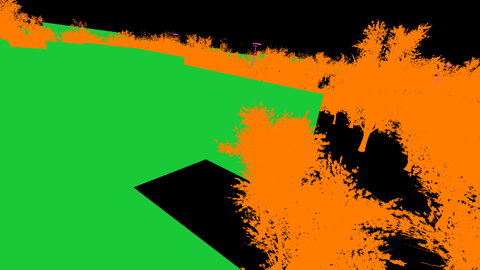}}\\[0.5mm]
\subfloat{\includegraphics[width=0.328\linewidth]{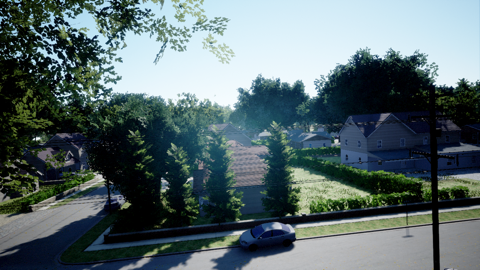}}\hfill
\subfloat{\includegraphics[width=0.328\linewidth]{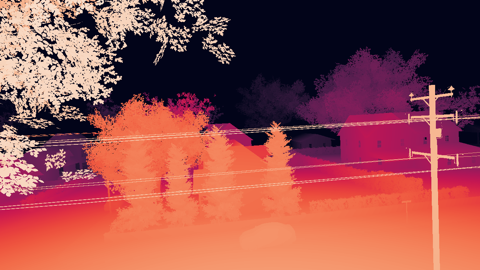}}\hfill
\subfloat{\includegraphics[width=0.328\linewidth]{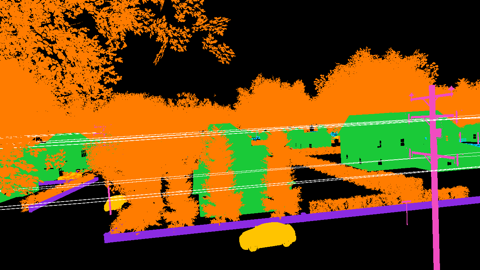}}\\[0.5mm]
\subfloat{\includegraphics[width=0.328\linewidth]{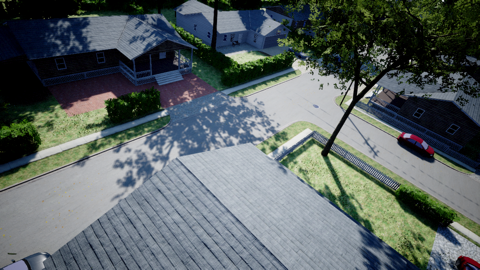}}\hfill
\subfloat{\includegraphics[width=0.328\linewidth]{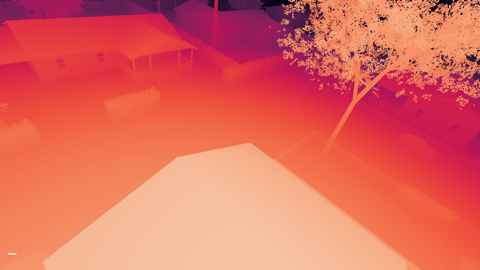}}\hfill
\subfloat{\includegraphics[width=0.328\linewidth]{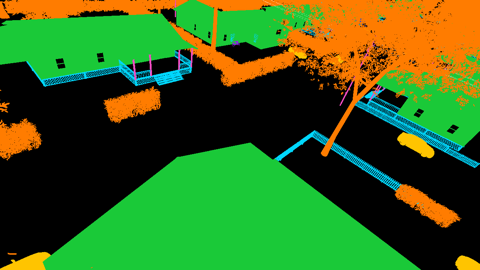}}\\[0.5mm]
\subfloat{\includegraphics[width=0.328\linewidth]{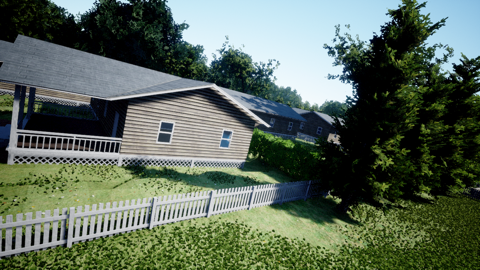}}\hfill
\subfloat{\includegraphics[width=0.328\linewidth]{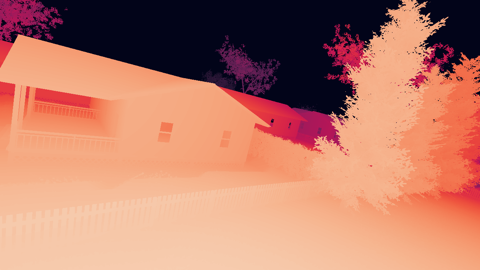}}\hfill
\subfloat{\includegraphics[width=0.328\linewidth]{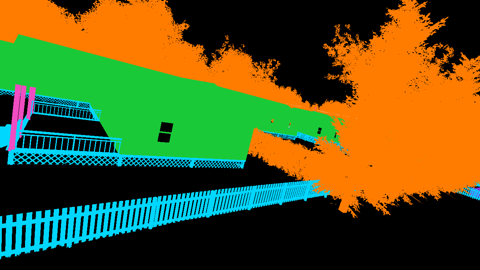}}

\caption{\textbf{Diverse perspectives in DDOS.} This selection highlights various aerial views from the DDOS dataset, with each frame presenting an RGB image, its depth map, and semantic segmentation. The imagery captures a range of features, from varied vegetation to complex architectural structures. Optical flow and surface normals, while part of the dataset, are not included in this visualization. Viewers are advised to examine these images digitally.}
\label{fig:ddos_additional_examples1}
\end{figure*}

\newpage

\section{Additional Examples}
\label{sec:additional_examples}

In this section, we present further examples from the DDOS dataset, as illustrated in \Cref{fig:ddos_additional_examples1,fig:ddos_additional_examples2}. These examples are specifically selected to highlight the dataset's diversity and the intricate details captured within. For clarity and emphasis on these finer aspects, the visualizations are confined to the RGB images, accompanied by their respective depth maps and semantic segmentations. Notably, \Cref{fig:ddos_additional_examples1} offers a glimpse into the diverse perspectives encompassed within DDOS. Conversely, \Cref{fig:ddos_additional_examples2} is dedicated to showcasing scenarios captured during low altitude flights in snowy conditions, underscoring the dataset's versatility and the challenging environments it encompasses.

DDOS, serves as a comprehensive aerial resource for the research community, particularly in the domains of depth estimation and segmentation. Its utility is especially evident in scenarios involving aerial perspectives, as encountered by drones, offering valuable insights for discerning thin structures within the visual field.

\end{document}